\documentclass[letterpaper]{article} 
\usepackage{aaai23-r2hcai}  
\usepackage{times}  
\usepackage{helvet}  
\usepackage{courier}  
\usepackage[hyphens]{url}  
\usepackage{graphicx} 
\usepackage{natbib}  
\usepackage{caption} 
\usepackage{algorithm}
\usepackage{algorithmic}

\usepackage{newfloat}
\usepackage{listings}
\DeclareCaptionStyle{ruled}{labelfont=normalfont,labelsep=colon,strut=off} 
\floatstyle{ruled}
\newfloat{listing}{tb}{lst}{}
\floatname{listing}{Listing}
\usepackage{fancyhdr}
\fancypagestyle{firstpagehf}
{
   \fancyhf{}
   \fancyhead[HC]{The AAAI 2023 Workshop on Representation Learning for Responsible Human-Centric AI (R$^2$HCAI)}
   \fancyfoot[C]{\thepage}
}

\title{A System for Human-AI collaboration for Online Customer Support}
\author{
    Debayan Banerjee\equalcontrib \quad
    Mathis Poser\equalcontrib \quad
    Christina Wiethof\equalcontrib \quad
    Varun Shankar Subramanian \quad
    Richard Paucar \quad
    Eva A. C. Bittner \quad
    Chris Biemann
}
\affiliations {
    Universität Hamburg, Hamburg, Germany \\
    \{debayan.banerjee,mathis.poser,christina.wiethof,eva.bittner,chris.biemann\}@uni-hamburg.de,\{varunshankar55,rfpaucar\}@gmail.com
}

\usepackage{bibentry}

\begin{document}
\thispagestyle{firstpagehf}
\maketitle

\begin{abstract}
AI enabled chat bots have recently been put to use to answer customer service queries, however it is a common feedback of users that bots lack a personal touch and are often unable to understand the real intent of the user's question. To this end, it is desirable to have human involvement in the customer servicing process. In this work, we present a system where a human support agent collaborates in real-time with an AI agent to satisfactorily answer customer queries. We describe the user interaction elements of the solution, along with the machine learning techniques involved in the AI agent.
\end{abstract}

\begin{figure*}[]
  \centering
  \includegraphics[width=0.9\textwidth]{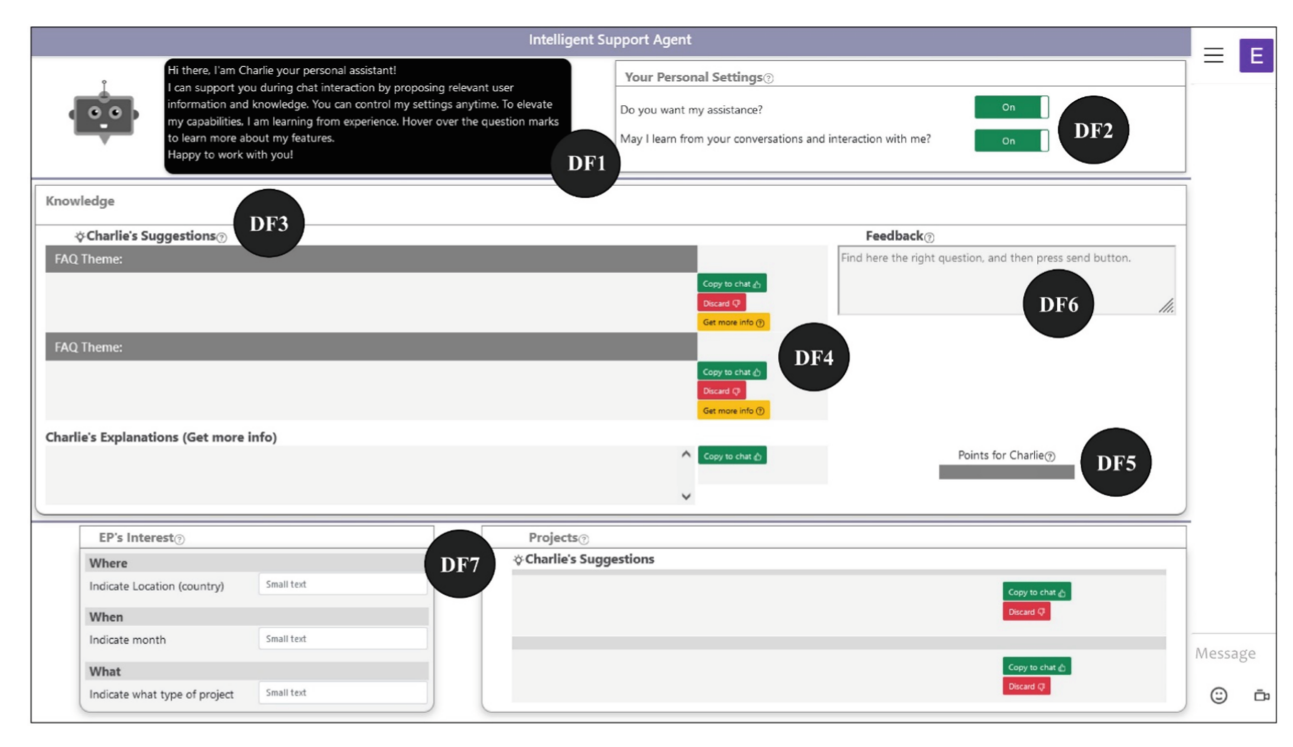}
  \caption{Screenshot of web based prototype}
  \label{ui1}
\end{figure*}

\section{Introduction}

In the pursuit of operational efficiency, companies across the globe have been deploying automation technology aided by Artificial Intelligence (AI) for Online Customer Support (OCS) use cases \footnote{\url{https://www.gartner.com/smarterwithgartner/4-key-tech-trends-in-customer-service-to-watch}}. With the explosive growth of social media usage, incoming customer queries have grown exponentially and to handle this growth, the use of proper technology is critical. Some estimates say that by the year 2025, 95\% of all customer interactions will be processed in some form by AI \footnote{\url{https://servion.com/blog/what-emerging-technologies-future-customer-experience/}}. However, AI in its present state is not advanced enough to completely replace human agents for most customer support scenarios. Additionally, the complete replacement of human workforce by AI is a topic of active ethical and political debate.  For these reasons the development of a hybrid working environment is required, where both human agents and AI agents can co-operate to satisfy OCS requirements. \\
In this work we briefly describe a web based user interface that allows a customer to interact with a human support agent, where the human agent receives helpful suggestions in parallel from an AI agent. In subsequent sections, we elaborate further on the machine learning techniques used for the AI agent.

Our present work is a part of a project which aims to find ways of integrating AI agents into customer support based workflows, with an aim of reducing workload of human agents. It is one of the primary goals of the project not to entirely replace the human agent with AI, and instead find productive  means of co-existence of the two.  As a part of this project, an international volunteer-driven organisation, which organises internships and projects for students across the globe was involved. In this organisation, prospective students participate in text based chat with human agents, and typically enquire about available opportunities and how to participate in them.  The human agents in turn use their domain expertise to provide the necessary information to the students. 

All the students and human agents involved were residents of Germany and hence the conversations were carried out in the German language. After collecting the conversations, an annotation phase was undertaken, where relevant utterances of the conversation were annotated with the corresponding FAQ IDs. When the conversations originally took place, there was no singular FAQ database in existence. For the purpose of this project, such a database was created. This made it possible to annotate the utterances with relevant FAQ IDs.

The goal of the dataset is to train an AI agent that can passively listen to the ongoing conversation and make relevant suggestions visible only to the human agent, not to the student. The human agent may then forward the suggested FAQ answer to the student, or decide not to do so if the quality of suggestion is poor. The eventual goal is for the human agent to spend less time looking for the right answer in a Knowledge Base, and instead offload this task to the AI agent. 

Later, a web UI was constructed, as described in the Web Interface section, that the human agent uses to interact with the student. The student is not aware of the UI's existence and is operating on a separate chat platform. The AI agent provides timely suggestions in this UI which is visible to the human agent. 

Our scenario differs from conventional Conversational Question Answering (CQA) or Interactive Information Retrieval (IIR) where the user interacts directly with the AI agent, and the AI agent is responsible for a response at each turn. In our case, the AI agent is in a passive listening role. It observes the ongoing conversation between two humans, and makes suggestions that are only visible to the human agent. Since the task of the AI agent is not just to suggest relevant FAQs but also to remain silent when no relevant FAQ is to be suggested, we evaluate both of these aspects in the evaluation section. 

The user interface  presented in this work has been published before \cite{10.1007/978-3-031-06516-3_11}. The machine learning techniques used to train the AI agent are yet to be published, and hence a larger focus in this work is on the AI training aspect.

\section{Web Interface} \label{webui}

The web-based frontend in Figure \ref{ui1} is labelled with certain design features (DF) to be explained shortly. The interface was implemented with Bootstrap and ReactJS while the backend API is hosted as a Python Flask app. The interface  greets humans agents with an avatar named Charlie that presents a brief usage
explanation (\textbf{DF1}). In addition, setting options for AI support and learning behavior are
provided (\textbf{DF2}). The integrated chat window is based on the open-source chat framework
Rocket Chat. The backend generates a ranked list of FAQ suggestions based on ML techniques to be described later.  In the frontend, two FAQ items - including
theme and accuracy in percent - with the highest agreement are displayed (\textbf{DF3}). The
discard buttons can be used to sequentially display four additional FAQ suggestions with
decreasing accuracy. The copy-to-chat buttons insert FAQ text into the input field of the chat window. Detailed information about a respective FAQ can be viewed via the get-more-info button (\textbf{DF4}). With a counter, points are added (copy-to-chat) or subtracted
(discard), if buttons are clicked (\textbf{DF5}). A feedback field allows entering search terms
to select and submit a FAQ that matches the interaction (\textbf{DF6}). Based on customers’
chat messages, exact keyword-based text matching is performed to automatically record
interests and suggest suitable projects from a database (\textbf{DF7}).

\begin{figure*}[]
  \centering
  \includegraphics[width=0.9\textwidth]{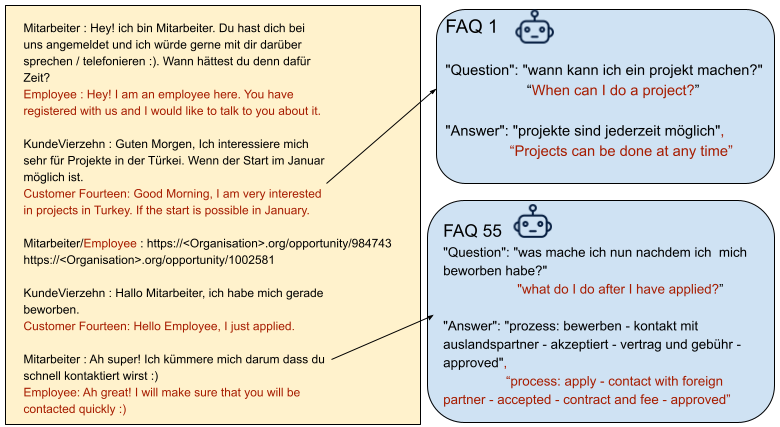}
  \caption{A sample conversation from the dataset with relevant corresponding FAQ annotation. The text in red is English translation of the conversation for the purpose of this paper, and not a part of the dataset.}
  
\end{figure*}

\section{Related Work}

The earliest dialogue systems, or chat-bots, were rule based \cite{eliza,parry} and subsequently corpus based chat-bots were developed \cite{https://doi.org/10.48550/arxiv.1512.05742} . In recent times neural chat-bots are frequently encountered in day to day customer support scenarios \cite{https://doi.org/10.48550/arxiv.2105.04387}.

 Recently, an interplay of human and AI collaboration in the process has been explored \cite{liu-etal-2021-role}. However current research in this area is focused on the AI bot being the first line of service, and only in the case of failures of the bot, a handover is initiated to a human agent, who plays a secondary role in the process. In contrast, our scenario makes the human agent the first line of support with the AI agent assisting in parallel.

To train chat-bots, conversational QA datasets such as the Ubuntu corpus \cite{ubuntu}, CoQA \cite{coqa}, DoQA \cite{doqa} and QuAC \cite{quac} have made progress in providing the community with rich grounds for conversational research. While CoQA relies on passages from broad domains such as children's stories and science to retrieve answers, QuAC relies on Wikipedia articles to create conversations and answers. DoQA on the other hand, focuses on three specific domains of cooking, travel and movies from stackexchange.com. In scope of how our dataset is modelled, it is most similar to DoQA, which is a domain specific conversational dataset which also requires retrieval of the correct FAQ from a database.
CoQA, DoQA and QuAC datasets are crowd-sourced and collected by the Wizard of Oz method. On the contrary, our dataset consists of genuine conversations between two humans whose sole purpose is to find the best internship possible for the student. During the conversations, neither of the parties were aware of the need to form an annotated dataset. Hence, our dataset has no artificial aspects in the flow of conversation.\\
The Dortmunder Chat Korpus \cite{dck} and The Verbmobil \cite{verbmobi} project provide German conversational corpus but they do not address the Question Answering or Information Retrieval domains.\\ 
Recently, the GermanQuAD and GermanDPR 
\cite{gquad} projects from DeepSet have enabled access to Transformer based models trained on the German text, which we make use of in our evaluation section, however the dataset they are based on is in the form of Questions and Answers, and not conversational in nature.\\

\section{Dataset Creation}

To train the AI agent, a conversational dataset had to be constructed. For this purpose, the conversations were carried out on the popular mobile application WhatsApp \footnote{https://play.google.com/store/apps/details?id=com.whatsapp}, where both the human agent and the student were on Whatsapp. The Web Interface described in the previous section was not included in this process. The conversations centered around topics such as how to register for a project, which projects are available in a given location, and whether there will be certifications available at the end etc. 
The chats were extracted using the export functionality of WhatsApp. The conversations have been collected over a period of two years, between 2018 and 2020. In some cases, an individual conversation may also span over a duration of several months, where the student and the human agent re-established contact after a gap of more than a few days. Such information is visible through the inclusion of the timestamp field in the dataset for each message that is exchanged.\\
Relevant consent for releasing their conversations was collected from the participating students and agents. Moreover, the identities of the participants and the organisation are pseudo-anonymised. Instead of the names of the participants, they are given a numerical name such as KundeSechsundzwanzig, which stands for Customer 26 in German. The human agent is represented by the term Mitarbeiter which stands for employee. 

A single human agent handled all the 26 conversations on WhatsApp over a period of time. When the conversations were carried out between 2018-2020, no single FAQ database existed at the organisation. The human agent instead used relevant domain expertise and experience within the organisation, and referred to a set of disjoint sources of information when the chats took place. Later in 2021, the human agent and a fellow domain expert colleague compiled a single FAQ database that covers most of the issues discussed in the conversations. Specific turns of the conversations were manually annotated with relevant FAQs by the human agent and then verified by the domain expert colleague.


\section{Dataset Analysis}

\begin{figure}[]
\includegraphics[width=0.5\textwidth]{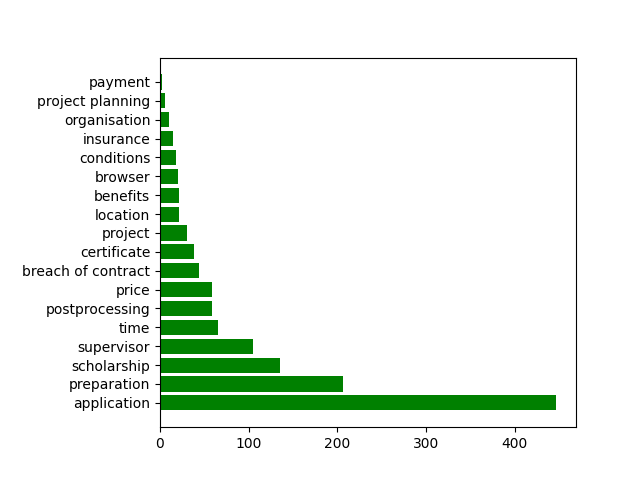}
\caption{Distribution of conversation topics in the dataset.}
\centering
\label{topicdist}
\end{figure}

\begin{figure}[]
\includegraphics[width=0.5\textwidth]{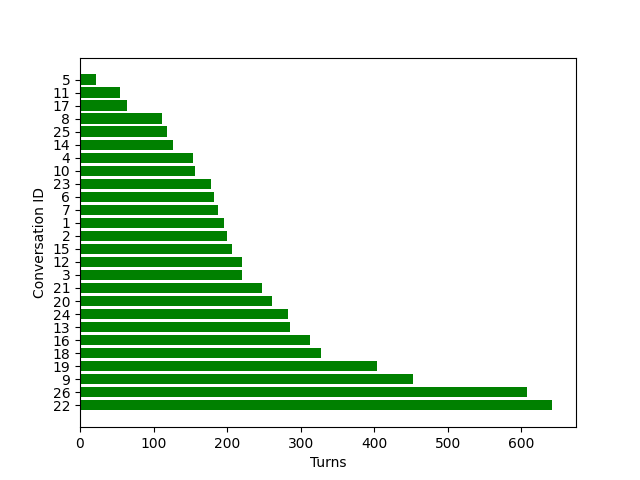}
\caption{The length of each conversation}
\centering
\label{convlength}
\end{figure}

\textbf{Chats and FAQs.} As depicted in Figure \ref{convlength} the 26 collected conversations vary in length ranging from 22 utterances to 607 utterances, with an average of 239 utterances per conversation.
The entire set of conversations consists of 6,219 utterances. 20.9 
\% of the utterances are annotated with the relevant FAQ ID. A significant portion of the dataset consists of chit-chat or other non-specific topics where no suggestion is supposed to be made by the AI agent to the human agent. \\
Since certain topics in the chat are discussed more often than others, as seen in Figure \ref{topicdist}, 
the distribution of relevant annotated FAQ IDs also is imbalanced
with FAQ ID 71 being the most frequent. FAQ 71 pertains to the procedure of registering online for projects. \\
We have split the dataset into train, dev and test splits in roughly 70:10:20 ratios. The train, dev and test splits have 17, 3 and 6 conversations, respectively, consisting of 3,693 , 891 and 1,635 utterances.\\

\section{Experimental Setting}

\subsection{Task Definition}

We define the task with the following inputs: current utterance $u_k$, the set of FAQs $F$, and the history of utterances so far $\{u_1,u_2,....,u_{k-1}\}$. The task for the model is to rank the correct FAQ item from $F$ to the top. If for a given utterance no FAQ is appropriate, the model must produce as the top-ranked output a special class that denotes absence of FAQ suggestion. We hereby call this class \texttt{no-suggestion}.

\subsection{Models}

As baselines we use the following settings:

\textbf{dumb} In this setting, the system produces 10 suggestions, with class \texttt{no-suggestion} at the top and FAQ IDs 1 to 9 as the subsequently ranked suggestions as output.

\textbf{random} In this setting, the system produces at random 10 classes as output without repetition. The output may contain one of the FAQ IDs or the \texttt{no-suggestion} class.

Additionally. we employed BM25 \cite{bm25} based text search ranking as a baseline method. In this method we searched the input query string against the FAQ database and used the ranked list of results. 


To produce strong performance, we employ Dense Passage Retrieval \cite{dpr} techniques . As a baseline, we use 
\textbf{fb-multiset-english}, which is a set of encoders \footnote{facebook/dpr-ctx\_encoder-multiset-base} that were pre-trained on English Natural Questions \cite{nq}, TriviaQA \cite{triviaqa}, WebQuestions \cite{webquestions}, and CuratedTREC \cite{curatedtrec}.   

Finally, we use pre-trained context and query encoders for the German language provided by DeepSet \footnote{\url{https://www.deepset.ai/germanquad}} and fine-tune them on our dataset for 100 epochs with a learning rate of 1e-05 with the Adam optimizer. We use random sampling for choosing negative examples during training. We choose the best performing model based on mrr@10 on the dev split. We used deepset-german encoders, which come comes from DeepSet and is trained on GermanQuAD \cite{gquad} dataset. \\

For query, we concatenate 4 consecutive utterances of conversation and consider it the input to the model. For context, we concatenate the question and answer for each FAQ and make the DPR model consider these as the passages database from which it has to rank the best possible FAQ.

\subsection{Evaluation Metrics}

As our metric, we choose the Mean Reciprocal Rank (MRR). For each query candidate, the model produces an MRR, which is the reciprocal of the position of the correct FAQ in the ranked list. We consider only the top 10 candidates, and hence, if the correct candidate is not in the top 10, we consider the MRR as 0. We compute the eventual MRR by taking a mean of the MRR of each query sample in the test set. \\
We evaluate separate MRRs for those utterances which have empty FAQ suggestions as gold annotation, and the ones which have non-empty FAQ gold suggestions. As explained before, the task of the AI agent is not just to recommend the right FAQ when needed, but it must also remain silent when no FAQ is suitable. We measure the ability of AI agent on both these tasks in Table \ref{tableresults}.

\subsection{Experimental Setup}

Since a large percentage of the utterances (79.1\%) belongs to the \texttt{no-suggestion} class we experiment with different mixture of \texttt{faq} classes and the \texttt{no-suggestion} class. During preparation of train and dev sets to be fed to the model, we calibrate the ratio of \texttt{no-suggestion} utterances differently as follows:\\
\textbf{mean} In this setting, we compute the mean of the frequency of the \texttt{faq} classes 
and include these many samples of randomly chosen \texttt{no-suggestion} utterances as input.\\
\textbf{highest-freq} In this setting, we find the most frequent \texttt{faq} class 
and include the same number of \texttt{no-suggestion} class samples.\\
\textbf{sum} In this setting, the number of samples of the utterances in \texttt{no-suggestion} class is equal to the sum of the number of utterances in all the \texttt{faq} classes combined. \\
\textbf{original} In this setting we consider all utterances as input which leads to roughly 80:20 class imbalance of \texttt{no-suggestion} class and the \texttt{faq} classes.\\
It must be noted that in all the above settings, we always include every \texttt{faq} class utterance. For input to the model we concatenate 4 consecutive utterances $\{u_{k-3},u_{k-2},u_{k-1},u_k\}$ for each utterance $u_k$.  When concatenating the utterances, we also append the sender name to the beginning of each utterance. 


\begin{table}[htb!]
\centering
\begin{tabular}{|p{28mm}|c|c|}
\hline
Model/Setting & \texttt{no-suggestion} & \texttt{faq} \\
\hline
dumb & 1.0 & 0.02\\
\hline
random & 0.04 & 0.06\\
\hline
BM25 & 0 & 0.27 \\
 
\hline
fb-multiset-english& &\\
\hspace{6mm}mean & 0.12 & 0.40\\
\hspace{6mm}highest-freq & 0.35 & 0.48 \\
\hspace{6mm}sum & 0.81 & 0.44 \\
\hspace{6mm}original & 0.96 & 0.33 \\
\hline
deepset-german & & \\
\hspace{6mm}mean & 0.12 & 0.58\\
\hspace{6mm}highest-freq & 0.42 & 0.57 \\
\hspace{6mm}sum & \textbf{0.84} & \textbf{0.50} \\
\hspace{6mm}original & 0.95 & 0.38 \\
\hline
\end{tabular}
\caption{MRR@10 values for different models and settings on test split of dataset}

\label{tableresults}
\end{table}

\section{Results}

We first analyse the baseline results from Table \ref{tableresults} : The \textbf{dumb} setting achieves perfect MRR in the \texttt{no-suggestion} category since in this setting the AI agent chooses 'silence' as the top ranked candidate for all turns. However it produces extremely poor results for turns that do require suggestions, since there is no intelligence or logic built in to his setting when fetching FAQ items. This also highlights why we need to evaluate our system on two different classes. If we had computed a singular MRR score for all turns, a model which remains silent all the time would score high accuracy. The \textbf{random} setting achieves poor performance in both categories. The \textbf{BM25} setting produces 0 MRR in \texttt{no-suggestion} class because there is no way to ask a text search method to not return any results. It always fetches some set of results, and in effect, is unable to produce silence as output. 

The Deep Passage Retrieval approaches using the \texttt{deepset-germandpr} set of models perform the best, which comes as no surprise since these encoders were pre-trained on German QA datasets, and further fine-tuned on our dataset. In comparison \texttt{fb-multiset-english} performs worse since the encoders are not aware of the German language. We find that among the different settings of varying proportions of the inclusion of \texttt{no-suggestion} class in the input, the \texttt{sum} setting produces a balanced performance in the two categories of \texttt{no-suggestion} and \texttt{faq}. 
 Another notable point in the table is the performance of the 
 \texttt{dumb} model which always produces \texttt{no-suggestion} as output hence achieving perfect MRR@10 of 1.0 in the relevant samples, but it produces the worst results in the \texttt{faq} classes, hence rendering it of little use to human agent. 
 We observe that as \texttt{no-suggestion} class performance improves, \texttt{faq} class performance drops. This brings forth interesting questions on how to calibrate the performance of the model to reach a sweet spot for the human agent. An MRR of 0.5 or greater for the \texttt{faq} classes means that the right FAQ is generally either in the first or in the second position, which is a positive contribution to lessen the human agent's workload, since most user interface implementations for our scenario would display the top 3 FAQs to human agent together. It is, however, more important for the \texttt{no-suggestion} MRR to be closer to 1.0, since the silence class being ranked second still produces suggestions that the human agent has to process, increasing noise for the human agent.

\section{Human Evaluation}

To evaluate the usability aspects of the prototype and its influence on the
task, we conducted interviews with 18 human agents after usage. Additionally, we inspected their
usage behavior via screen recordings to supplement the qualitative results. Overall,
human agents indicated that they would continue to use the prototype and highlighted that it is
particularly helpful for agents who do not have much experience in handling customers.
 During customer interactions, agents sent on average 16 (SD: 5; Median: 14) messages
during the customer interaction. 17 agents used the FAQ answer suggestions via the copy-to-chat-button at least three times. On average, agents edited two (SD: 2; Median: 2) of the
suggested responses in the input field before sending them. 

Overall, an average of six (SD: 2.5; Median: 7)
suggestions were used, whereby the detailed version via get-more-info button (Mean:
3.7; SD: 2.6; Median: 4.5) was used more frequently than the short version (Mean: 2.6;
SD: 2.4; Median: 2). To receive alternative FAQ answer suggestions, the discard-button
was clicked on average 15 times (SD: 10.8; Median: 15). The display of two suggestions
and the option for additional explanatory information via the get-more-info-button were
perceived as helpful \textit{“so that you can think in which direction you might go”} (agent1). Agents
experienced relief through displayed suggestions and the majority saved time making
decisions, especially by using the copy-to-chat-button: \textit{“ I just had to copy them,
which affected the speed”} (agent14).  16 agents utilized the feedback function
on average four times, while nine people successfully provided feedback. However, agents
expressed the need for an adaptation of the feedback function, as it was unclear. 
Concerning the recommendation of projects, the pressure to recall knowledge or search
in parallel to the customer interaction was reduced as relevant information was presented.
Thereby, it \textit{“took out the uncomfortable part of working with such a consultation,
which is looking up stuff ”} (agent16)

\section{Limitations}

The current solution suffers from the following limitations: 1) The web interface was developed for internal evaluation purposes and is not available for general public use. 2) The collection of the dataset suffers from class imbalance and bias issues, since only a single person was involved in collecting the conversations. 3) The feedback function of the UI did not work as expected by the human agents. The human agents expected the feedback regarding wrong suggestions to be immediately learnt by the system, however during the evaluation phase we did not re-train our models, or perform on-line learning from the provided feedback. 

\section{Conclusion and Future Work}

In this work we present a web interface for demonstrating hybrid human-AI collaborative system that can handle customer support queries. We show through machine based and human based evaluations, that with the limited and imbalanced data we collected, we found appropriate methods to train an AI agent that is able to provide appropriate assistance to its human counterpart, which is the goal of our research.

For future work, we wish to implement active on-line learning from the human agent's usage of the feedback feature in the UI. We would also like to collect a larger and more balanced dataset for future iterations of the AI agent.

\section{Acknolwedgements}
The research was financed with funding provided by the German Federal Ministry of Education and Research and the European Social Fund under the "Future of work"
program (INSTANT, 02L18A111).

\bibliography{aaai23}

\end{document}